\documentclass{sigkddExp}

\usepackage{amsmath}
\usepackage{graphicx}
\usepackage{color}
\usepackage{soul}
\usepackage{subcaption}

\begin{document}
%

\title{Predicting Individual Physiologically Acceptable States for Discharge from a Pediatric Intensive Care Unit}

\author{
\alignauthor Cameron Carlin, Long Van Ho, David Ledbetter, Melissa Aczon, Randall Wetzel \\[5pt]
       \affaddr{The Laura P. and Leland K. Whittier Virtual Pediatric Intensive Care Unit}\\
       \affaddr{Children's Hospital Los Angeles}\\
       \affaddr{4650 Sunset Blvd Los Angeles, CA}\\[1.5pt]
       \email{\{ccarlin, loho, dledbetter, maczon, rwetzel\}@chla.usc.edu } \\
}
\maketitle
    
\section*{ABSTRACT}

\textbf{Objective:}  Predict patient-specific vitals deemed medically acceptable for discharge from a pediatric intensive care unit (ICU).\\[2pt]  

\noindent{\textbf{Design:} 
The means of each patient's heart rate, systolic and diastolic blood pressure measurements between their medical and physical discharge from the ICU were computed as a proxy for their physiologically acceptable state space (PASS) for successful ICU discharge.  These individual PASS values were compared via root mean squared error (rMSE) to  population age-normal vitals, a polynomial regression through the PASS values of a Pediatric ICU (PICU) population and predictions from two recurrent neural network models designed to predict personalized PASS within the first twelve hours following ICU admission.}\\[2pt]  

\noindent{\textbf{Setting:} 
PICU at Children's Hospital Los Angeles (CHLA).}\\[2pt]

\noindent{\textbf{Patients:} 
6,899 PICU episodes (5,464 patients) collected between 2009 and 2016.}\\[2pt]

\noindent{\textbf{Interventions:} None.}\\[2pt]

\noindent{\textbf{Measurements:}
Each episode data contained 375 variables representing vitals, labs, interventions, and drugs. They also included a time indicator for PICU medical discharge and physical discharge.}\\[2pt]

\noindent{\textbf{Main Results:}
The rMSEs between individual PASS values and population age-normals (heart rate: 25.9 bpm, systolic blood pressure: 13.4 mmHg, diastolic blood pressure: 13.0 mmHg) were larger than the rMSEs corresponding to the polynomial regression (heart rate: 19.1 bpm, systolic blood pressure: 12.3 mmHg, diastolic blood pressure: 10.8 mmHg).  The rMSEs from the best performing RNN model were the lowest (heart rate: 16.4 bpm; systolic blood pressure: 9.9 mmHg, diastolic blood pressure: 9.0 mmHg).}\\[2pt]

\noindent{\textbf{Conclusion:}
PICU patients are a unique subset of the general population, and general age-normal vitals may not be suitable as target values indicating physiologic stability at discharge.  Age-normal vitals that were specifically derived from the medical-to-physical discharge window of ICU patients may be more appropriate targets for `acceptable' physiologic state for critical care patients.  Going beyond simple age bins, an RNN model can provide more personalized target values.

\section{INTRODUCTION}

\begin{figure*}
\centering 
\includegraphics[scale=0.35]{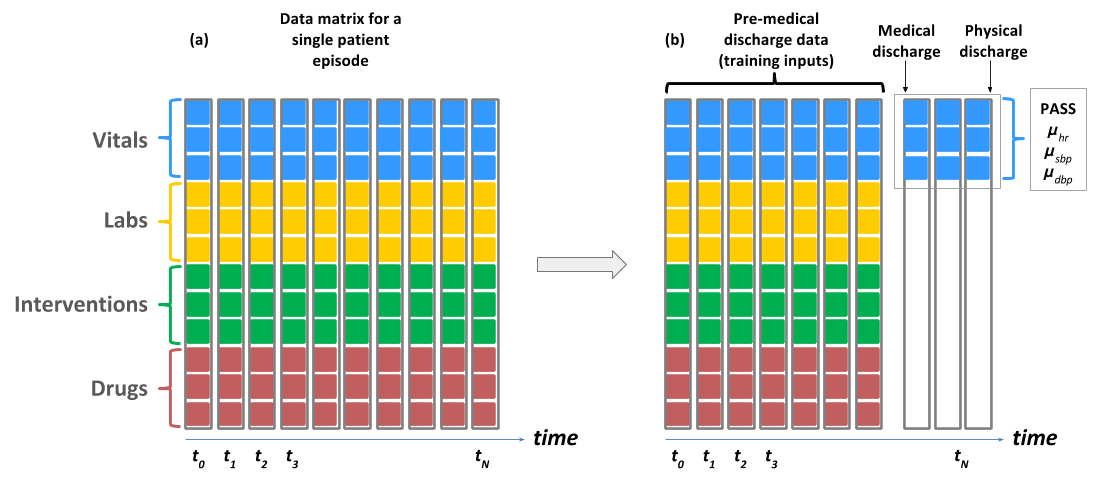}
\caption{(a) After pre-processing, data for a single patient episode are in a matrix format.  A single row of data contains values (actual and imputed measurements) from a single variable. A column of data comprises all measurements at one time point. Reproduced with permission from \protect\cite{aczon2017dynamic,ho2017dependence}.  (b)  The means of a surviving patient's vitals between medical and physical discharge from the ICU define this patient's ICU-stable or Physiologically Acceptable State Space (ICU-PASS).  Data from the pre-medical discharge window are used to predict individual ICU-PASS values.}
\label{fig:data_matrix} 
\end{figure*} 

Patients in an intensive care unit (ICU) undergo high-fre-quency monitoring and are treated to achieve and maintain homeostatic stability approximating health 
\cite{yeh1984validation}.  Implicit in that goal is resuscitating the patient towards normal homeostasis, often defined by the clinical team's experience and expertise. This includes restoring a patient's vitals towards a medically accepted range of values.  In practice, these conceptual `target' values are often implied but not explicitly defined.  Measures of health and risk of mortality rely on the deviance of routine physiologic signs from expected normal values \cite{pollack1997pediatric,slater2003pim2,duncan2006pediatric}.  In this paper, we investigated the question of what is ‘normal’ for discharge from a Pediatric ICU (PICU).
\\[2pt]

Data describing age-normal vital signs and their implications in a clinical setting \cite{fleming2011normal,knudson1976maximal,nugent2006emergency} focus on the expectations for a healthy individual. However, the physiologic status of patients in an ICU may differ from those of the normal population. ICU admission criteria metrics \cite{chalmers2011severity,lockrem1989recommendations,dawson1993admission,nates2016icu} include assessments of a patient's condition to understand the severity of their mental and physical health. The Society of Critical Care Medicine guidelines \cite{nates2016icu} consider the patient's condition, their potential for improvements from interventions, and the duration of their condition as criteria for ICU admission.\\[2pt]

Because of the distinction between ICU normal populations, the physiologic state considered stable or acceptable for ICU discharge may not necessarily be the same as that of a healthy individual \cite{pronovost2003improving}. The purpose of this study was two-fold:  quantify the difference between acceptable ICU-dis-charge and population age-normal vitals; and develop a machine learning model that predicts these acceptable vitals for individual patients. From a pediatric critical care database, an age dependent ICU-discharge physiologically acceptable state was derived using polynomial regression and compared to published age-normal values. A machine learning model that predicts an individual patient's physiologically acceptable values for discharge from the PICU was also developed, which may enable personalized targets for patient care management.

\section{MATERIALS AND METHODS}

\subsection{Data Description}

The data were extracted from anonymized observational clinical data collected in Electronic Medical Records (EMR, Cerner) in the Pediatric Intensive Care Unit (PICU) of Children's Hospital Los Angeles between 2009 and 2016. The CHLA IRB reviewed the study protocol and waived the need for consent or need for IRB approval. Each episode had time series measurements for 375 variables representing vitals, laboratory results, interventions and drugs. Pre-processing techniques described in previous work \cite{aczon2017dynamic,ho2017dependence} converted this EMR into matrix format (feature vector), illustrated in Figure 1(a), to make it amenable to machine learning algorithm development. The figure depicts a single patient's physiologic state space, represented with a 375-dimensional vector at each time point.\\[2pt] 

To define the physiologic state considered acceptable for discharge from the PICU, vital signs prior to discharge from the PICU were analyzed. The PICU routinely recorded when patients were medically ready for discharge as determined by clinician judgement.  This was followed by some period, during which the patient remained monitored, before actual physical discharge from the PICU (.25 quartile/median/.75 quartile were 7hrs/9hrs/28hrs).  Data collected between the medical discharge time and the physical discharge time from the ICU were considered to represent the physiologically acceptable state space (PASS) for each child. Only survivors with at least 3 measurements of heart rate (HR), systolic blood pressure (SBP) and diastolic blood pressure (DBP) within this window were included. Finally, only those episodes lasting at least 12 hours after ICU admission were included. These constraints resulted in a database of 6899 uniquely identified episodes (length of stay .25 quartile/median/.75 quartile were 35hrs/61hrs/120hrs) corresponding to 5464 distinct patients.\\[2pt]

These episodes were randomly split into training, validation and test sets prior to analysis. To prevent possible leakage, this division was done by patients such that all episodes from a single patient belonged to only one of the three sets. Approximately sixty percent of patients were placed in the training set (4320 episodes, 3313 patients), twenty percent in the validation set (1419 episodes, 1097 patients), and twenty percent in the test set (1160 episodes, 1054 patients).  

\subsection{Defining and Predicting Patient-Specific ICU-PASS}

For each of the 6,899 surviving episodes with a medical-physical discharge window, the means of heart rate ($\mu_{hr}$), systolic blood pressure ($\mu_{sbp}$), and diastolic blood pressure ($\mu_{dbp}$) within this period, as illustrated by Figure 1(b), were computed and defined the episode's ICU Physiologically Acceptable State Space (ICU-PASS). Heart rate, systolic blood pressure, and diastolic blood pressure were selected as vitals critical to determining physiologic stability \cite{yeh1984validation}.  PASS vital signs from individual episodes in the training set were used to develop a polynomial regression model that depends only on age.  A fifth order polynomial for each vital was chosen from the inflection point of the bias-variance trade-off when iterating through different polynomial orders. This regression model was used to predict individual PASS values and compared to published age-normals.\\[2pt]

\begin{figure*}
\centering 
\includegraphics[scale=0.3]{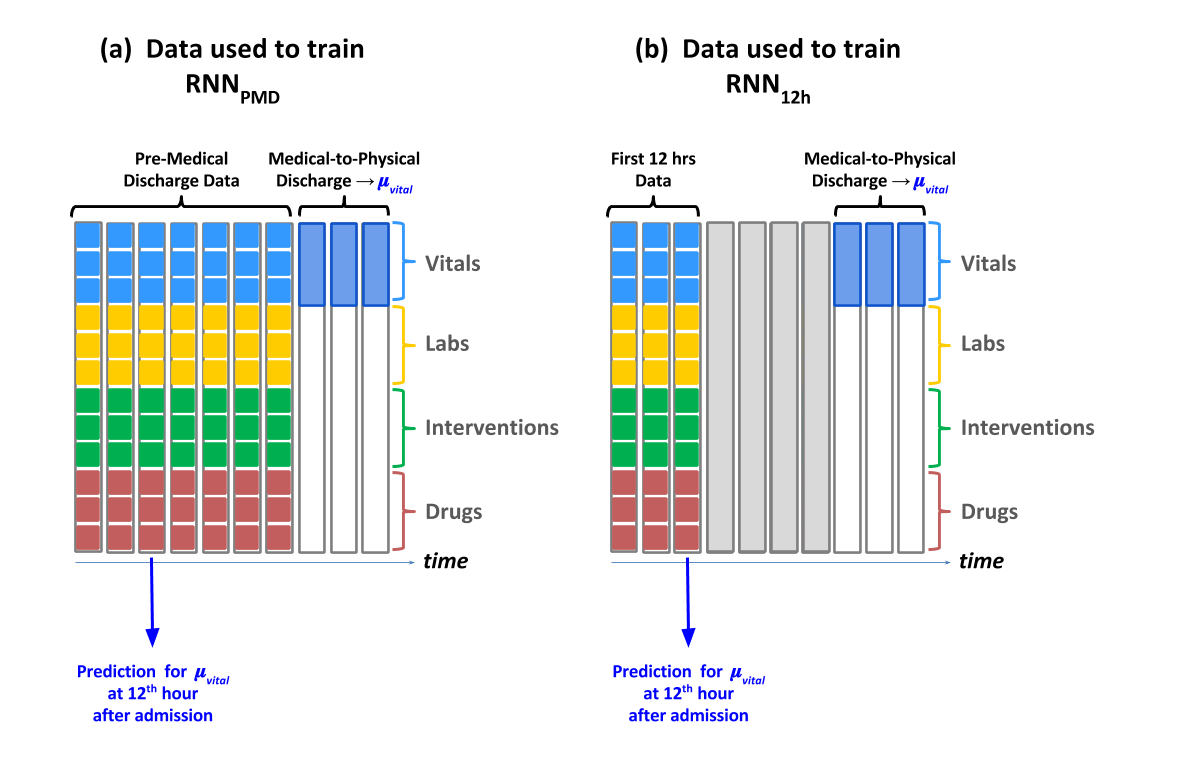} 
\caption{(a) During training, RNN$_{\text{PMD}}$ minimizes errors over all pre-medical discharge time points. (b) RNN$_{\text{12h}}$ minimizes errors over the first 12 hours only. Both models predict the means of heart rate ($\mu_{hr}$), systolic blood pressure ($\mu_{sbp}$), and diastolic blood pressure ($\mu_{dbp}$) computed from the medical-to-physical discharge window.  During assessment, predictions generated at the $12^{th}$ hour following ICU admission are used for comparing model errors.}
\label{fig:data_matrices_with_vital} 
\end{figure*} 

Other factors besides age may affect ICU-PASS vitals. Recurrent Neural Networks (RNNs) were used to learn these factors. Designed with a feedback loop, RNNs can sequentially ingest time series data, integrate them and learn temporal relationships \cite{aczon2017dynamic, hochreiter1997long, greff2015lstm, jozefowicz2015empirical}. Medical applications on which RNNs have been used successfully include a time-varying severity of illness score \cite{aczon2017dynamic}, early detection of critical decompensation in children  \cite{shah20162}, onset of heart failure \cite{choi2017using}, de-identification of patient notes \cite{dernoncourt2017identification}, and disease diagnosis from EMR \cite{razavian2015temporal}.\\[2pt]

\begin{table*}
\centering
\caption{Comparing 12th hour rMSE and MAE between RNN vital predictions, ICU Acceptable predictions, and Age-Normal predictions to patient-specific true `acceptable' values. For each vital, the p-value of ANOVA F-Tests between models were less than 0.01.}
\label{tab:HR_rMSEs}
\begin{tabular}{|l|cc|cc|cc|}
\hline
& \multicolumn{6}{c|}{\textbf{Target Vital}} \\ \hline
  & \multicolumn{2}{c|}{\textbf{Heart Rate (bpm)}} & \multicolumn{2}{c|}{\textbf{Systolic BP (mm Hg)}} & \multicolumn{2}{c|}{\textbf{Diastolic BP (mm Hg)}} \\ 
  & \textbf{rMSE} & \textbf{MAE} & \textbf{rMSE} & \textbf{MAE} & \textbf{rMSE} & \textbf{MAE} \\
\hline
\textbf{Age-Normal} & 25.9 & 21.2 & 13.4  & 10.7 & 13.0 & 10.4\\
\textbf{Regression}  & 19.1 & 15.3 & 12.3  & 9.7 & 10.8 & 8.5 \\
\hline
\textbf{RNN$_{\text{12h}}$}  & 18.1 & 14.5 & 10.1 & 8.0 & 9.1 & 7.2 \\
\textbf{RNN$_{\text{PMD}}$}   & \textbf{16.4} & \textbf{12.9} & \textbf{9.9} & \textbf{7.8} & \textbf{9.0} & \textbf{7.0} \\
\hline
\end{tabular}
\end{table*}

\begin{table*}
\centering
\caption{Comparison of heart rate performance (errors in bpm) parsed by primary diagnosis.}
\label{tab:diagnosis_bin_results}
\begin{tabular}{|l|cc|cc|cc|cc|}
\hline
& \multicolumn{8}{c|}{\textbf{Primary Diagnosis}} \\ \hline
       & \multicolumn{2}{c|}{\textbf{Spine Curve Disorders}} & \multicolumn{2}{c|}{\textbf{ARDS}} & \multicolumn{2}{c|}{\textbf{Brain Neoplasm}} & \multicolumn{2}{c|}{\textbf{Sepsis}}  \\
              & \multicolumn{2}{c|}{\textbf{N=118}} & \multicolumn{2}{c|}{\textbf{N=78}} & \multicolumn{2}{c|}{\textbf{N=70}} & \multicolumn{2}{c|}{\textbf{N=68}}  \\
              & \textbf{rMSE}                  & \textbf{MAE}                  & \textbf{rMSE}            & \textbf{MAE}             & \textbf{rMSE}            & \textbf{MAE}            & \textbf{rMSE}        & \textbf{MAE}               \\ \hline
\textbf{Age Norm}                 & 35.2                          & 31.7                         & 27.4                    & 21.8                    & 19.5                    & 15.9                   & 27.5                & 23.3                             \\
\textbf{Regression}                  & 22.2                          & 19.0                         & 18.6                    & 14.4                    & 21.8                    & 17.9                   & 19.0                & 14.2                                  \\ \hline
\textbf{RNN$_{\text{12h}}$}           & 18.0                          & 14.6                         & 18.7                    & 13.9                    & 21.2                    & 17.7                   & 17.5                & 13.4                                 \\
\textbf{RNN$_{\text{PMD}}$} & \textbf{17.6}                 & \textbf{14.0}                & \textbf{17.5}           & \textbf{13.1}           & \textbf{16.6}           & \textbf{13.9}          & \textbf{17.0}       & \textbf{12.8}           \\ \hline     
\end{tabular}
\end{table*}

Two RNN models, both of which used Hochreiter's Long Short-Term Memory (LSTM) architecture \cite{hochreiter1997long}, were developed to predict each individual patient's ICU-PASS values. The models used data from the pre-medical discharge period shown in Figure 1(b) to predict ICU-PASS values, $\mu_{hr},\, \mu_{sbp},\,\mu_{dbp}$. The first model (RNN$_{\text{PMD}}$) was trained on all the time points before medical discharge of each episode in the training set, while the second (RNN$_{\text{12h}}$) trained only on data through the first 12 hours of those same episodes.  Figure \ref{fig:data_matrices_with_vital} illustrates the partitioning of each episode data matrix into columns used for model training and columns used to compute actual ICU-PASS values. The validation set was used to optimize RNN hyper-parameters\cite{james2013introduction}.

\subsection{Method of Assessment}

The 1160 episodes in the test set were used to validate the predictions made by the models. The RNN predictions made at the 12th hour following ICU admission, values determined from the polynomial regression, and published age-normals were compared to each individual child's actual ICU-PASS values. Two standard error metrics were used for each vital: root mean squared error (rMSE):
\begin{equation*}
\frac{1}{N_{test}} \sqrt{\sum_{i=1}^{N_{test}}  |\mu_{v}(P_i) - \hat\mu_{v}(P_i(t=12^{th} \text{hr}))|^2},
\end{equation*}
and mean absolute error (MAE):  
\begin{equation*}
\frac{1}{N_{test}} \sum_{i=1}^{N_{test}} |\mu_{v}(P_i) - \hat\mu_{v}(P_i(t=12^{th} \text{hr}))|,
\end{equation*}
where $N_{test}$ denotes the number of patient episodes in the test set. The notation $\mu_{v}(P_i)$ represents the actual ICU-PASS value for a particular vital ($\mu_{hr}$, $\mu_{sbp}$, or $\mu_{dbp}$) of the $i^{th}$ patient episode in the test set, which is constant in time,  while $\hat \mu_{v}(P_i)$ denotes model prediction for it at the 12th hour following ICU admission.  When comparing the various models, we computed p-values of an ANOVA test \cite{dunn1974applied} between baseline age-normal and each of the other three models. \\[2pt]

Error aggregation over all test set encounters, as indicated the in above equations, grants a perspective of model validity over the PICU population as a whole. Performance was also measured on stratified population subsets, including age ranges associated with medically accepted age-normal vitals, primary diagnoses, and PIM score quartiles.

\begin{figure*}
\centering
\begin{subfigure}{\textwidth}
  \centering
  \includegraphics[width=.85\linewidth]{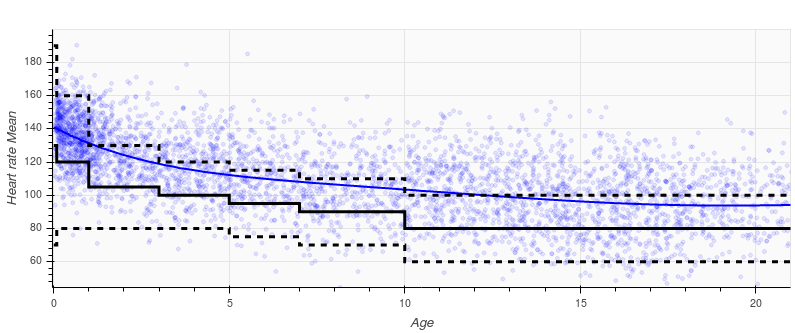}
  \caption{Heart Rate}
  \label{fig:hr_fit}
\end{subfigure}
\begin{subfigure}{\textwidth}
  \centering
  \includegraphics[width=.85\linewidth]{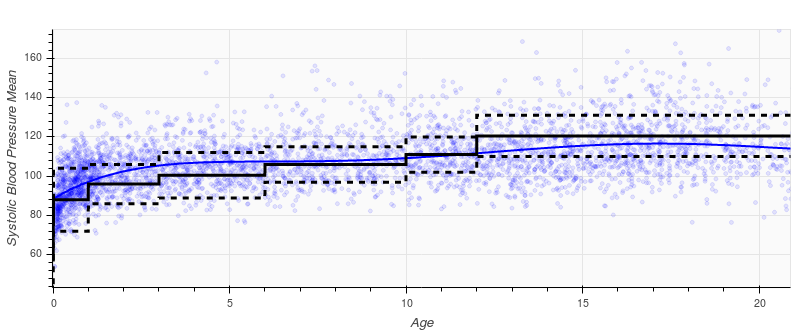}
  \caption{ Systolic Blood Pressure}
  \label{fig:sbp_fit}
\end{subfigure}
\begin{subfigure}{\textwidth}
  \centering
  \includegraphics[width=.85\linewidth]{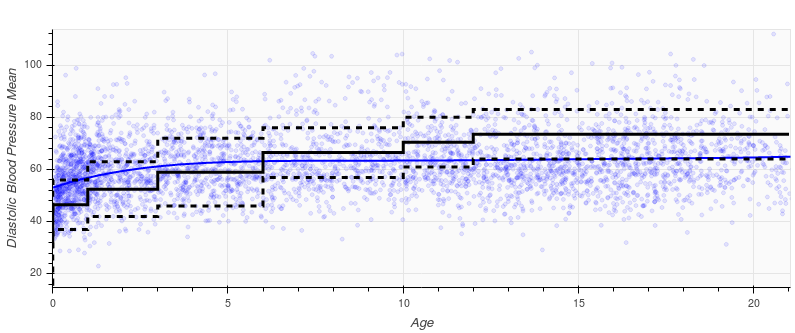}
  \caption{Diastolic Blood Pressure}
  \label{fig:dbp_fit}
\end{subfigure}
\begin{subfigure}{\textwidth}
  \centering
  \includegraphics[width=.2\linewidth]{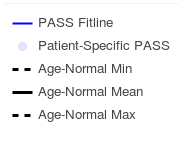}
  \label{fig:fit_legend}
\end{subfigure}
\caption{PASS points (blue dots), Polynomial Regression (solid blue) and published Age-Normal values (black) for heart rate, systolic blood pressure, and diastolic blood pressure.}
\label{fig:hr_sbp_dbp_fit}
\end{figure*}

\section{RESULTS}

The ICU-PASS heart rate, systolic and diastolic blood pressure values from individual patient episodes in the training set were plotted as a function of age in Figure \ref{fig:hr_sbp_dbp_fit}.  Evident from these plots is the wide scatter of ICU-PASS values within any particular age.  Superimposed on the ICU-PASS values are two age dependent functions: the polynomial regression model (blue line) and previously published acceptable age-normal ranges for vitals (dashed black lines) \cite{simel2011approach,pomerance1997nelson,falkner2004fourth} with their corresponding means (solid black line).  Compared to the regression curve, age-normal heart rate values are uniformly and significantly lower by 10 to 25 bpm.  A similar comparison shows age-normal systolic blood pressure values to be lower by up to 9mmHg for children younger than 6 years old, but higher by 4 to 9 mmHg for those older than 12 years. Also relative to the regression curve, diastolic blood pressure values are lower by 3 to 11 mmHg for patients younger than 6 years old, but higher by 2 to 9 mmHg for those older.\\[2pt]

Table \ref{tab:HR_rMSEs} displays the two error metrics, parsed separately for heart rate, systolic and diastolic blood pressures, from the four models considered. The predictions from the polynomial regression equation outperformed the mean age-normals from literature.  Both RNN models, $\mbox{RNN}_{\text{PMD}}$ and RNN$_{\text{12h}}$, performed more accurately in predicting each individual patient's true ICU-PASS vitals.  RNN$_{\text{PMD}}$ performed most accurately across all vitals and error metrics. 

\subsection{Results within Specific Diagnoses}

Results were partitioned by primary diagnosis; as an example, Table \ref{tab:diagnosis_bin_results} displays the results for heart rate for four frequently occurring diagnoses.  For episodes with Spine Curve Disorders, ARDS or Sepsis, the regression model predictions incurred smaller errors than literature age-normals. RNN$_{\text{PMD}}$ remained the best performing model within each of the diagnoses. RNN$_{\text{12h}}$ performed similarly to $\mbox{RNN}_{\text{PMD}}$ for Spine Curve Disorders and Sepsis patients, yet did not show significant improvement from the regression model for ARDS and Brain Neoplasm patients. For Brain Neoplasm patients, RNN$_{\text{12h}}$ also performed worse than the age-normal values; only RNN$_{\text{PMD}}$ was able to outperform the age-normal values for these patients.

\subsection{Results By Age-Normal Bins}

The errors were also partitioned according to the age bins in the literature.  As shown in Table \ref{table:age_bin_results}, RNN$_{\text{PMD}}$ had the lowest rMSE and MAE across all ages except for the 0-1 month old bin. The true PASS heart rate of patients who were younger than a month were most closely approximated by the polynomial regression model.  RNN$_{\text{12h}}$ showed improvement from the age-normal and regression model for all patients older than one year. At all age bins, the literature age-normals had greater MAEs and rMSEs than the regression model.

\subsection{Results by PIM2 Score}

Table 4 displays the errors partitioned by PIM2 quartiles. The regression model fit showed significant improvements over mean age-normals across all PIM2 quartiles. RNN$_{\text{12h}}$ showed comparable
performance to the regression model. RNN$_{\text{PMD}}$ showed the best performance across all quartiles.

\begin{table*}
\centering
\caption{Comparison of heart rate performance (errors in bpm) parsed by age bin.}
\label{table:age_bin_results}
\begin{tabular}{|l|ccccccc}
\hline
                      & \multicolumn{7}{c|}{\textbf{Age Bin MAEs}}                                                                                                                                                       \\ \hline
                      & \textbf{0-1 Mos.} & \textbf{1-11 Mos.} & \textbf{1-2 Years} & \textbf{3-4 Years} & \textbf{5-6 Years} & \textbf{7-9 Years} & \multicolumn{1}{c|}{\textbf{10+ Years}} \\
                       & \textbf{N = 10} & \textbf{N = 169} & \textbf{N = 186} & \textbf{N = 91} & \textbf{N = 59} & \textbf{N = 142} & \multicolumn{1}{c|}{\textbf{N = 503}} \\ \hline
\textbf{Age-Normal} & 19.8               & 18.5                & 21.2                & 18.7                & 20.7                & 20.7                & \multicolumn{1}{c|}{22.3}                \\
\textbf{Regression}    & \textbf{10.0}    & 13.0                & 12.5                & 15.9                & 15.7                & 14.0                & \multicolumn{1}{c|}{17.4}                \\ \hline
\textbf{RNN$_{\text{12h}}$}    & 20.8               & 14.1                & 11.9                & 14.6                & 14.1                & 12.6                & \multicolumn{1}{c|}{16.0}                \\
\textbf{RNN$_{\text{PMD}}$}    & 11.9               & \textbf{12.3}     & \textbf{11.4}     & \textbf{14.4}     & \textbf{13.4}     & \textbf{12.3}     & \multicolumn{1}{c|}{\textbf{13.6}}     \\ \hline
\end{tabular}
\end{table*}

\begin{table*}
\centering
\caption{Comparison of heart rate performance (errors in bpm) parsed by PIM2 score quartiles, from lowest to highest severity of illness.}
\label{table:pim2}
\begin{tabular}{|l|cc|cc|cc|cc|} \hline
& \multicolumn{8}{c|}{\textbf{PIM 2 Score Quartiles}} \\ \hline
& \multicolumn{2}{c|}{\textbf{First Quartile}} & \multicolumn{2}{c|}{\textbf{Second Quartile}} & \multicolumn{2}{c|}{\textbf{Third Quartile}} & \multicolumn{2}{c|}{\textbf{Fourth Quartile}} \\ 
              & \textbf{rMSE}                  & \textbf{MAE}                  & \textbf{rMSE}            & \textbf{MAE}             & \textbf{rMSE}            & \textbf{MAE}            & \textbf{rMSE}        & \textbf{MAE}               \\ \hline
\textbf{Age-normal} & 25.4 & 20.8 & 27.9 & 23.1 & 24.0 & 19.6 & 25.5 & 20.6 \\
\textbf{Regression}   & 19.6 & 16.0 & 19.1 & 15.6 & 18.4 & 14.9 & 19.0 & 14.8 \\
\textbf{RNN$_{\text{12h}}$}     & 18.4 & 15.0 & 17.7 & 14.2 & 17.5 & 14.2 & 18.8 & 14.7 \\
\textbf{RNN$_{\text{PMD}}$}     & \textbf{16.0} & \textbf{13.0} & \textbf{16.5} & \textbf{13.0} & \textbf{15.4} & \textbf{12.2} & \textbf{17.6} & \textbf{13.5} \\ \hline
\end{tabular}
\end{table*}

\section{DISCUSSION}

We have described the physiologic state, defined by heart rate and blood pressure that characterizes when a child is well enough, in clinical judgement, to be discharged from the ICU.  It is clear that this state is not identical to published healthy norms. The ability to estimate a patient's physiologically acceptable status for discharge from the ICU within twelve hours of their admission may have several uses in developing further machine learning approaches to clinical data and the clinical management of critically ill children. First, we undertook this project to enable us to compare values obtained during an ICU stay to what would be `acceptable' for a given patient.  We intended to understand deviation from this acceptable level, and the volatility of the observations as a measure of homeostatic instability.  This required being able to posit an individual patient's own `acceptable' values. Secondly, being able to posit a child's individual physiologically acceptable state space suggests the potential for developing personalized physiological targets for patient management. In the ICU, one generally considers age-normal values as a guide for individual patient therapy. The wide and generalized ranges of age norms make them less useful for predicting individualized values. Further, it is clear that published values differed from what was considered physiologically acceptable for discharge in this study.\\[2pt] 

The two RNN models developed, RNN$_\text{PMD}$ and RNN$_\text{12h}$, exhibited significant improvement over published values and values predicted by the regression model, suggesting that the RNN models may enable more personalized care. The distinction between the two models is important. While RNN$_\text{PMD}$ was trained to learn patient trajectories from admission to medical discharge (a window spanning more than 12 hours), RNN$_\text{12h}$ learned trajectories only from the first 12 hours following ICU admission.  In all error comparisons, RNN$_\text{PMD}$ outperformed RNN$_\text{12h}$. Both RNN models can generate predictions in real time as new clinical measurements become available, but the analyses presented here were based on their 12th hour predictions of an individual patient's PASS values determined in the pre-discharge period. \\[2pt]

Machine learning approaches to EMR medical data are becoming more common and have been increasingly reported \cite{aczon2017dynamic,shah20162,choi2017using,dernoncourt2017identification,razavian2015temporal}.  We have demonstrated that an advanced deep learning methodology using the rich data in the EMR, rather than simply using a relationship between age and vital signs, more accurately predicted individual patient's PASS for discharge from the PICU. Better understanding individual deviation from discharge acceptable values has potential for wider machine learning and AI applications in medical practice. \\[2pt] 

Model accuracy is important over the entire PICU population and over specific subpopulations. This is especially true if a proposed model is to replace existing baselines for guiding patient-specific target vitals. Examining the results by diagnosis enabled more patient-specific analysis.  Patients with more generalized systemic diseases, such as those with ARDS and Sepsis \cite{force2012acute,nguyen2006severe}, are expected to have vitals with a greater deviance from what would be considered `healthy' than those with localized disease processes such as Brain Neoplasms.  These are consistent with the results in Table \ref{tab:diagnosis_bin_results}, where the heart rate errors between age-normal and true ICU-PASS values were greater (rMSE $>$ 27 bpm) for patients with ARDS, Sepsis or Spine Curve Disorders than those with Brain Neoplasms (rMSE $<$ 20 bpm). In contrast, the errors corresponding to the regression model and RNN models had smaller variations across the diagnostic categories. The only category in which RNN$_\text{PMD}$ was not the best performing model was the 0-1 month age bin, which was the least populated bin containing 36 patients in the training set and 10 patients in the test set.   

\section{CONCLUSION}

ICU-PASS predictions from our age-dependent regression and RNN models provide more accurate references for acceptable discharge vitals in pediatric intensive care populations than that of age-normal references for a healthy population. We have shown substantial improvement in predicting discharge physiologic status using historical PICU data. Likewise, RNN predictions based on rich medical knowledge derived from a patient's treatments and vital signs, not just age, yielded more precise predictions of their specific `medically-acceptable' ICU vitals. Future work will aim to expand the predicted ICU-PASS vitals to include a wider perspective of patient physiology, allowing broader comparisons to existing pediatric patient-state measures such as PIM 2 and PRISM III. Furthermore, we hope to expand these results to not only understand the PASS expectations, but to better understand the patient's stability and variance during the medical to physical discharge window. Understanding expectations of patient stability and vitality as they are acceptable to depart from the ICU will allow a broader understanding of the methodology and temporal relationship between a patient and their condition throughout their PICU encounter.

\section{Acknowledgements}
This work was funded by a grant from the Laura P. and Leland K. Whittier Foundation.

\bibliographystyle{abbrv}
\bibliography{bibliography}  

\section{Appendix}
Tables below contain supplementary results for systolic and diastolic blood pressure.

\begin{table*}[]
\centering
\caption{Comparison of systolic blood pressure performance (errors in mmHg) parsed by age bin.}
\label{my-label}
\resizebox{1.0\textwidth}{!}{%
\begin{tabular}{|l|cc|cc|cc|cc|cc|cc|} \hline
\textbf{}           & \multicolumn{12}{c|}{\textbf{Systolic Blood Pressure}}                                                                                                                                                                                                                                                                     \\ \hline
\textbf{Age Bin}    & \multicolumn{2}{c|}{\textbf{0-12 Months}} & \multicolumn{2}{c|}{\textbf{1-2 Years}} & \multicolumn{2}{c|}{\textbf{3-5 Years}} & \multicolumn{2}{c|}{\textbf{6-9 Years}} & \multicolumn{2}{c|}{\textbf{10-11 Years}} & \multicolumn{2}{c|}{\textbf{$>$12 Years}} \\
\textbf{Metric}     & \textbf{rMSE}                  & \textbf{MAE}                  & \textbf{rMSE}                & \textbf{MAE}                & \textbf{rMSE}          & \textbf{MAE}         & \textbf{rMSE}         & \textbf{MAE}          & \textbf{rMSE}          & \textbf{MAE}           & \textbf{rMSE}          & \textbf{MAE}         \\ \hline
\textbf{Age-normal} & 13.8                          & 10.5                         & 11.8                        & 9.4                        & 13.4                  & 10.6                & 11.8                 & 9.4                  & 10.7                  & 8.9                   & 14.4                  & 11.9                \\
\textbf{PASS Fit}   & 11.7                          & 9.1                          & 10.2                        & 8.2                        & 12.9                  & 10.4                & 11.8                 & 9.4                  & 10.7                  & 9.0                   & 13.3                  & 10.4                \\ \hline
\textbf{RNN$_{12h}$}     & 9.8                           & 7.7                          & 9.0                         & 7.2                        & 11.1                  & 8.9                 & 9.7                  & 7.8                  & 8.6                   & 6.8                   & 10.8                  & 8.6                 \\
\textbf{RNN$_{PMD}$}    & \textbf{9.6}                  & \textbf{7.6}                 & \textbf{8.9}                & \textbf{7.1}               & \textbf{10.7}         & \textbf{8.5}        & \textbf{9.5}         & \textbf{7.5}         & \textbf{8.4}          & \textbf{6.7}          & \textbf{10.7}         & \textbf{8.4}   \\ \hline    
\end{tabular}}
\end{table*}

\begin{table*}[]
\centering
\caption{Comparison of systolic blood pressure performance (errors in mmHg) parsed by primary diagnosis.}
\label{my-label}

\begin{tabular}{|l|cc|cc|cc|cc|} \hline
\textbf{}           & \multicolumn{8}{c|}{\textbf{Systolic Blood Pressure}}                                                                                                                       \\ \hline
         & \multicolumn{2}{c|}{\textbf{Spine Curve Disorders}} & \multicolumn{2}{c|}{\textbf{ARDS}} & \multicolumn{2}{c|}{\textbf{Brain Neoplasm}} & \multicolumn{2}{c|}{\textbf{Sepsis}} \\
     & \multicolumn{2}{c|}{\textbf{N = 118}}                   & \multicolumn{2}{c|}{\textbf{N = 78}}   & \multicolumn{2}{c|}{\textbf{N = 70}}             & \multicolumn{2}{c|}{\textbf{N = 68}}     \\
\textbf{Metric}     & \textbf{rMSE}            & \textbf{MAE}            & \textbf{rMSE}   & \textbf{MAE}    & \textbf{rMSE}        & \textbf{MAE}         & \textbf{rMSE}     & \textbf{MAE}    \\ \hline
\textbf{Age-normal} & 13.8                    & 11.4                   & 11.9           & 9.5            & 12.6                & 10.6                & 14.5             & 12.3           \\
\textbf{PASS Fit}   & 11.9                    & 10.0                   & 10.4           & 8.4            & 10.8                & 8.9                 & 13.0             & 10.9           \\ \hline
\textbf{RNN$_{12h}$}     & 10.4                    & 8.5                    & 9.0            & 7.5            & 9.5                 & 8.0                 & \textbf{11.5}    & 9.3            \\
\textbf{RNN$_{PMD}$}    & \textbf{9.8}            & \textbf{7.9}           & \textbf{8.9}   & \textbf{7.1}   & \textbf{9.3}        & \textbf{7.9}        & 11.6             & \textbf{9.2}  \\ \hline
\end{tabular}
\end{table*}

\begin{table*}[]
\centering
\caption{Comparison of systolic blood pressure performance (errors in mmHg) parsed by PIM2 score quartiles, from lowest to highest severity of illness.}
\label{my-label}
\begin{tabular}{|l|cc|cc|cc|cc|} \hline
\textbf{}           & \multicolumn{8}{c|}{\textbf{Systolic Blood Pressure}}                                                                                                                  \\ \hline
\textbf{Quartile}   & \multicolumn{2}{c|}{\textbf{(0, 0.25)}} & \multicolumn{2}{c|}{\textbf{(0.25, 0.5)}} & \multicolumn{2}{c|}{\textbf{(0.5, 0.75)}} & \multicolumn{2}{c|}{\textbf{(0.75, 1)}} \\
\textbf{Metric}     & \textbf{rMSE}      & \textbf{MAE}      & \textbf{rMSE}       & \textbf{MAE}       & \textbf{rMSE}       & \textbf{MAE}       & \textbf{rMSE}      & \textbf{MAE}      \\ \hline
\textbf{Age-normal} & 12.6              & 10.2             & 13.6               & 10.5              & 12.8               & 10.3              & 13.9              & 11.3             \\
\textbf{PASS Fit}   & 11.2              & 8.9              & 12.8               & 9.9               & 11.9               & 9.5               & 12.6              & 10.2             \\ \hline
\textbf{RNN$_{12h}$}     & 9.6               & 7.6              & \textbf{10.5}      & 8.4               & 9.7                & 7.5               & 10.6              & 8.6              \\
\textbf{RNN$_{PMD}$}    & \textbf{9.2}      & \textbf{7.2}     & 10.6               & \textbf{8.3}      & \textbf{9.5}       & \textbf{7.4}      & \textbf{10.5}     & \textbf{8.4}    \\ \hline
\end{tabular}
\end{table*}

\begin{table*}[]
\centering
\caption{Comparison of diastolic blood pressure performance (errors in mmHg) parsed by age bin.}
\label{my-label}
\resizebox{1.0\textwidth}{!}{%
\begin{tabular}{|l|cc|cc|cc|cc|cc|cc|} \hline
\textbf{}           & \multicolumn{12}{c|}{\textbf{Diastolic Blood Pressure}}                        \\ \hline
    & \multicolumn{2}{c|}{\textbf{0-12 Months}} & \multicolumn{2}{c|}{\textbf{1-2 Years}} & \multicolumn{2}{c|}{\textbf{3-5 Years}} & \multicolumn{2}{c|}{\textbf{6-9 Years}} & \multicolumn{2}{c|}{\textbf{10-11 Years}} & \multicolumn{2}{c|}{\textbf{$>$12 Years}} \\
    & \textbf{rMSE}                  & \textbf{MAE}                  & \textbf{rMSE}                & \textbf{MAE}                & \textbf{rMSE}         & \textbf{MAE}          & \textbf{rMSE}         & \textbf{MAE}          & \textbf{rMSE}          & \textbf{MAE}           & \textbf{rMSE}         & \textbf{MAE}          \\ \hline
\textbf{Age-normal} & 13.3                          & 10.2                         & 12.0                        & 9.6                        & 11.5                 & 9.2                  & 10.6                 & 8.8                  & 10.9                  & 9.4                   & 14.1                 & 11.5                 \\
\textbf{PASS Fit}   & 9.8                           & 7.7                          & 9.2                         & 7.3                        & 11.9                 & 9.7                  & 10.4                 & 8.4                  & 9.4                   & 7.3                   & 11.1                 & 8.8                  \\ \hline
\textbf{RNN$_{12h}$}     & \textbf{8.4}                  & 6.6                          & \textbf{8.4}                         & 6.7                        & \textbf{9.9}         & \textbf{7.9}         & 9.3                  & 7.3                  & \textbf{7.9}                   & 6.0                   & 9.6                  & 7.5                  \\
\textbf{RNN$_{PMD}$}    & 8.5                           & \textbf{6.5}                 & \textbf{8.4}                & \textbf{6.6}               & \textbf{9.9}                  & 8.0                  & \textbf{9.1}         & \textbf{7.2}         & \textbf{7.9}          & \textbf{5.9}          & \textbf{9.3}         & \textbf{7.3}   \\ \hline     
\end{tabular}}
\end{table*}

\begin{table*}[]
\centering
\caption{Comparison of diastolic blood pressure performance (errors in mmHg) parsed by primary diagnosis.}
\label{my-label}
\begin{tabular}{|l|cc|cc|cc|cc|} \hline
\textbf{}           & \multicolumn{8}{c|}{\textbf{Diastolic Blood Pressure}}                                                                                                                      \\ \hline
        & \multicolumn{2}{c|}{\textbf{Spine Curve Disorders}} & \multicolumn{2}{c|}{\textbf{ARDS}} & \multicolumn{2}{c|}{\textbf{Brain Neoplasm}} & \multicolumn{2}{c|}{\textbf{Sepsis}} \\
     & \multicolumn{2}{c|}{\textbf{N = 118}}                   & \multicolumn{2}{c|}{\textbf{N = 78}}   & \multicolumn{2}{c|}{\textbf{N = 70}}             & \multicolumn{2}{c|}{\textbf{N = 68}}     \\
    & \textbf{rMSE}            & \textbf{MAE}            & \textbf{rMSE}   & \textbf{MAE}    & \textbf{rMSE}        & \textbf{MAE}         & \textbf{rMSE}    & \textbf{MAE}     \\ \hline
\textbf{Age-normal} & 13.6                    & 11.7                   & 10.1           & 8.5            & 13.1                & 10.9                & 13.1            & 10.6            \\
\textbf{PASS Fit}   & 9.2                     & 7.5                    & 8.7            & 7.0            & 10.0                & 8.3                 & 10.6            & 8.7             \\ \hline
\textbf{RNN$_{12h}$}     & 8.1                     & 6.3                    & \textbf{8.2}   & \textbf{6.5}   & 8.7                 & 6.7                 & \textbf{9.4}    & \textbf{7.7}    \\
\textbf{RNN$_{PMD}$}    & \textbf{8.0}            & \textbf{6.2}           & 8.8            & 7.0            & \textbf{8.5}        & \textbf{6.6}        & 10.1            & 8.0           \\ \hline 
\end{tabular}
\end{table*}

\begin{table*}[]
\centering
\caption{Comparison of diastolic blood pressure performance (errors in mmHg) parsed by PIM2 score quartiles, from lowest to highest severity of illness.}
\label{my-label}
\begin{tabular}{|l|cc|cc|cc|cc|} \hline
\textbf{}           & \multicolumn{8}{c|}{\textbf{Diastolic Blood Pressure}}                                                                                                                 \\ \hline
& \multicolumn{2}{c|}{\textbf{(0, 0.25)}} & \multicolumn{2}{c|}{\textbf{(0.25, 0.5)}} & \multicolumn{2}{c|}{\textbf{(0.5, 0.75)}} & \multicolumn{2}{c|}{\textbf{(0.75, 1)}} \\
     & \textbf{rMSE}      & \textbf{MAE}      & \textbf{rMSE}       & \textbf{MAE}       & \textbf{rMSE}       & \textbf{MAE}       & \textbf{rMSE}      & \textbf{MAE}      \\ \hline
\textbf{Age-normal} & 13.0              & 10.7             & 13.2               & 10.5              & 12.1               & 9.7               & 12.5              & 10.1             \\
\textbf{PASS Fit}   & 9.9               & 7.8              & 11.0               & 8.6               & 10.5               & 8.4               & 10.7              & 8.4              \\ \hline
\textbf{RNN$_{12h}$}     & 8.5               & 6.6              & 9.5                & \textbf{7.4}               & 8.8                & 7.0               & \textbf{9.6}      & 7.6              \\
\textbf{RNN$_{PMD}$}    & \textbf{8.2}      & \textbf{6.5}     & \textbf{9.4}       & \textbf{7.4}      & \textbf{8.6}       & \textbf{6.8}      & 9.7               & \textbf{7.5}    \\ \hline
\end{tabular}
\end{table*}

\end{document}